\documentclass[10pt,journal,compsoc,twoside]{IEEEtran}
\usepackage{amsmath,amsfonts,amssymb}
\usepackage{array}
\usepackage[caption=false,font=normalsize,labelfont=sf,textfont=sf]{subfig}
\usepackage{textcomp}
\usepackage{url}
\usepackage{verbatim}
\usepackage{graphicx}
\usepackage{cite}
\usepackage{booktabs}
\usepackage[section]{placeins}

\begin{document}

\title{A Linear Fractional Transformation Model and Calibration Method for Light Field Camera}

\author{ Zhong Chen, \IEEEmembership{Member, IEEE}, Changfeng Chen
\thanks{
    Zhong Chen and Changfeng Chen, are all with School of Mechanical and Automotive Engineering, South China University of Technology, Guangzhou, 510640 Chna.
}
\thanks{Manuscript received xxx xx, 20xx; revised xx xx, 20xx.}
}

\markboth{IEEE TRANSACTIONS ON PATTERN ANALYSIS AND MACHINE INTELLIGENCE,~Vol.~xx, No.~x, xxx~20xx}%
{Chen \MakeLowercase{\textit{et al.}}: A Linear Fractional Transformation Model and Calibration Method for Light Field Camera}

\IEEEpubid{0000--0000/00\$00.00~\copyright~2021 IEEE}

\maketitle

\begin{abstract}
Accurate intrinsic calibration is a crucial yet challenging prerequisite for 3D reconstruction using light field cameras. Existing calibration models typically analyze the main lens and micro lens array (MLA) in a coupled manner, resulting in high complexity and a large number of parameters. In this paper, we propose a linear fractional transformation (LFT) model that introduces a single parameter $\alpha$ to decouple the imaging processes of the main lens and the MLA. A dedicated matrix $\mathbf{H}_\alpha$ is designed to characterize the MLA projection, enabling the main lens and the MLA to be calibrated independently. The proposed calibration method consists of an analytical least-squares solution for $\mathbf{H}_\alpha$, followed by joint nonlinear refinement of all intrinsic parameters. Experimental results on both physical datasets and simulated data demonstrate that the proposed method achieves a mean translation error of $2.1\%$, outperforming the state-of-the-art, while maintaining sub-pixel reprojection accuracy. The complete codebase, including a light field simulator based on the proposed model, is openly available to the research community.
\end{abstract}


\section{Introduction}
\IEEEPARstart{L}ight field cameras \cite{2006Ng,2009Lumsdaine} utilize a Micro-lens Array (MLA) introduced between the objective lens and the sensor to simultaneously record the spatio-angular information. Owing to their low cost and high integration, they are widely applied in various fields such as medical imaging \cite{lu2025physics}, 3D reconstruction \cite{zhang20213d}, and microscopy \cite{lu2025physics,guo2019fourier}. To support the diverse applications of light field cameras, accurately calibrating them to obtain internal geometric parameters is crucial \cite{zhang2018generic}. Given the compact structure of MLA light field cameras, obtaining accurate intrinsic and extrinsic parameters through direct measurement is challenging. Therefore, researchers often calibrate the internal and external parameters of light field cameras by combining a camera model with feature data. However, the presence of the micro-lens array results in a complex camera model \cite{labussiere2020blur}, and the resulting inherent strong coupling between camera parameters significantly increases the difficulty of calibration.

Since the advent of light field cameras, researchers have proposed various calibration methods and models, most of which are based on the thin-lens pinhole model \cite{labussiere2020blur,2017Nousias,2017Noury} or the two-plane model \cite{zhang2018generic,2013Dansereau}. However, these approaches still suffer from several unresolved issues. First, methods based on the two-plane model require the construction of sub-aperture images\cite{2013Dansereau}, while those based on the thin-lens pinhole model involve complex formulations with redundant parameters\cite{labussiere2020blur}; consequently, the overall calibration process is rather cumbersome. Second, the parameters in many models are coupled , necessitating joint optimization \cite{2020Liu}, which increases the degrees of freedom in the solution space and leads to unstable calibration results. Third, some methods introduce parameters solely for computational simplification \cite{2017Nousias}, lacking physical meaning or practical utility as auxiliary criteria for system configuration assessment.

To address the strong coupling and complexity inherent in MLA light field cameras, we establish the Linear Fractional Transformation (LFT) model under the thin-lens and pinhole assumption. We then propose the LFT matrix $\mathbf{H}_\alpha$ to characterize the MLA, thereby achieving a crucial decoupling between the main lens and the MLA processes. Our main contributions are summarized as follows:

\begin{enumerate}
    \item We introduce the LFT matrix $H_\alpha$, derived from the geometric optical path analysis, to model the MLA effect in light field cameras.
    \item We develop a complete LFT model by integrating the MLA characteristics with Zhang's calibration framework.
    \item We propose a novel calibration method for light field cameras, including an analytic solution for the LFT matrix and a full-model nonlinear optimization. Source code is available on the authors' homepage.
\end{enumerate}

\section{related work}
The calibration of light field cameras is addressed in several publications, which broadly categorize the cameras into two types: unfocused light field cameras \cite{2006Ng} (or plenoptic camera 1.0) and focused light field cameras \cite{2009Lumsdaine} (or plenoptic camera 2.0).

Dansereau et al. \cite{2013Dansereau} established the first comprehensive mathematical model for the unfocused light field camera and introduced a method for decoding raw light field images into rays. Their model, based on reconstructed sub-aperture images (SAIs). It was later refined by Zhou et al. \cite{zhou2019two}, who enabled a physically based calibration method. Yunsu et al. \cite{2017Bok} developed a geometric projection model that directly utilizes raw images to estimate both intrinsic and extrinsic parameters, thereby avoiding the need for reconstruction. They also proposed a method for detecting line features in raw images, moving beyond conventional corner detection algorithms. Similarly, Zhao et al. \cite{2020Zhao} leverage the concept of plenoptic discs \cite{o2018calibrating} to develop a camera model for plenoptic camera 1.0. However, their applicability is largely confined to the unfocused configuration and cannot be readily extended to focused or multi-focus plenoptic camera systems.

With the arrival of commercial plenoptic camera 2.0 \cite{2009Lumsdaine}, growing research interest has been directed toward focused plenoptic camera and generalized models. Johannsen et al. \cite{2013johannsen} first formulated a general reprojection model in terms of the focused plenoptic camera and accounting for distortion. They also defined the virtual depth of light field and it was developed by Wang et al. \cite{wang2018virtual}. Noury et al. \cite{2017Noury} introduced a more complete geometrical model than the previous works. They proposed a new way to detect the corners in raw images. However, those methods take all micro-lenses as the same, which is not suitable for multi-focus configuration.

Nousias et al.\cite{2017Nousias} respectively considered different types of micro-lenses and proposed a focused light field camera model that can adapt to multi-focus configurations. Liu et al. \cite{2020Liu} proposed the stepwise calibration of light field cameras, based on the plenoptic disc feature \cite{o2018calibrating}. They adopt the algorithm proposed by Fleischmann et al. \cite{fleischmann2014lens} to distinguish the different types of micro-lenses. These methods take different configuration of micro-lenses as different intrinsic parameters, leading to a large number of parameters and makes the calibration more complex.

For generalized models applicable to both multi-focus and unfocused light field cameras, the state-of-the-art (SOTA) method was proposed by Labussière et al. \cite{2022mathieu}. They introduced a blur-aware light field camera model and calibration method that leverages blur information to improve calibration performance. However, although the blur scaling factor is theoretically constant, it must be calibrated per main-lens focusing distance, which adds complexity. Earlier, Zhang et al. \cite{zhang2018generic} presented a generic multi-projection-center (MPC) model. However, their method requires SAIs for calibration, which prevents direct simulation of raw images.

In summary, prior works either lack generality across different light field camera types, or suffer from high complexity caused by parameter coupling and the need for per-configuration calibration. A common underlying issue is the tight coupling between the main lens and the MLA in existing models. To address this, we propose a decoupled model based on linear fractional transformation, which simplifies calibration while preserving accuracy and universality.

\section{Light Field Model}

We consider the focused light field camera (plenoptic 2.0), in which the MLA is placed at a distance from the sensor that differs from the main lens focal length. This configuration is the most widely adopted in practice as it balances angular and spatial resolution \cite{2009Lumsdaine}. Depending on whether the MLA plane lies in front of or behind the main lens focal plane, the focused configuration subdivides into the Galilean and Keplerian types, respectively. As demonstrated by Nousias et al.\ \cite{2017Nousias}, the two types are equivalent and can be described by a common projection matrix. The analysis below begins with the MLA, the defining optical element that distinguishes light field cameras from conventional cameras, and then integrates the main lens and distortion to construct the complete projection model. Table~\ref{tab:notation} summarizes the notation used throughout this chapter.

\begin{table}[htbp]
\centering
\caption{Summary of notation used in the LFT model. Unless otherwise noted, all quantities are expressed in the camera coordinate system $\{\mathbf{C}\}$ or its pixel-space equivalent.}
\label{tab:notation}
\begin{tabular}{ll}
\toprule
\textbf{Symbol} & \textbf{Definition} \\
\midrule
\multicolumn{2}{l}{\textit{Camera geometry}} \\
\midrule
$(X_c, Y_c, Z_c)^T$ & Object point \\
$(X'_c, Y'_c, Z'_c)^T$ & Virtual image point formed by the main lens \\
$F$ & Focal length of the main lens \\
$d_c$ & Distance from the sensor plane to the main lens \\
$d_m$ & Distance from the MLA plane to the main lens \\
$(l_x, l_y)^T$ & Microlens center \\
\midrule
\multicolumn{2}{l}{\textit{Pixel-space quantities}} \\
\midrule
$s_x, s_y$ & Pixel scale factors (horizontal, vertical) \\
$(u, v)^T$ & Final image point \\
$m_x, m_y$ & Microlens center: $m_x \triangleq l_x s_x,\; m_y \triangleq l_y s_y$ \\
$q_x, q_y$ & Virtual image point: $q_x \triangleq X'_c s_x,\; q_y \triangleq Y'_c s_y$ \\
\midrule
\multicolumn{2}{l}{\textit{MLA grid}} \\
\midrule
$(s_{\text{base}}, t_{\text{base}})^T$ & Reference microlens center \\
$r$ & Microlens diameter \\
$\mathbf{R}(\theta)$ & In-plane rotation of the MLA \\
\midrule
\multicolumn{2}{l}{\textit{Distortion and projection}} \\
\midrule
$\alpha$ & LFT coefficient: $\alpha = (Z'_c - d_c)/(Z'_c - d_m)$ \\
$k_1, k_2$ & Radial distortion coefficients \\
$t_1, t_2$ & Tangential distortion coefficients \\
$\mathbf{H}_\alpha$ & MLA projection matrix \\
$\mathbf{H}$ & Complete light field projection matrix \\
\bottomrule
\end{tabular}
\end{table}

\subsection{Projection Model of the MLA}
\begin{figure}[ht]
\centering
\subfloat[]{\includegraphics[width=0.45\linewidth]{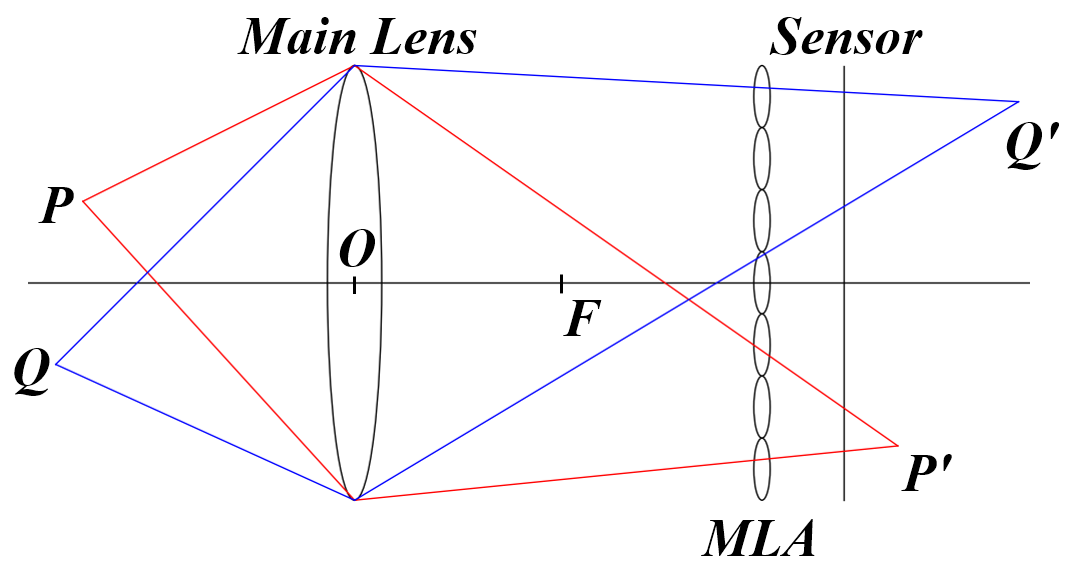}%
\label{fig:Mainlensproj}}
\hfil
\subfloat[]{\includegraphics[width=0.35\linewidth]{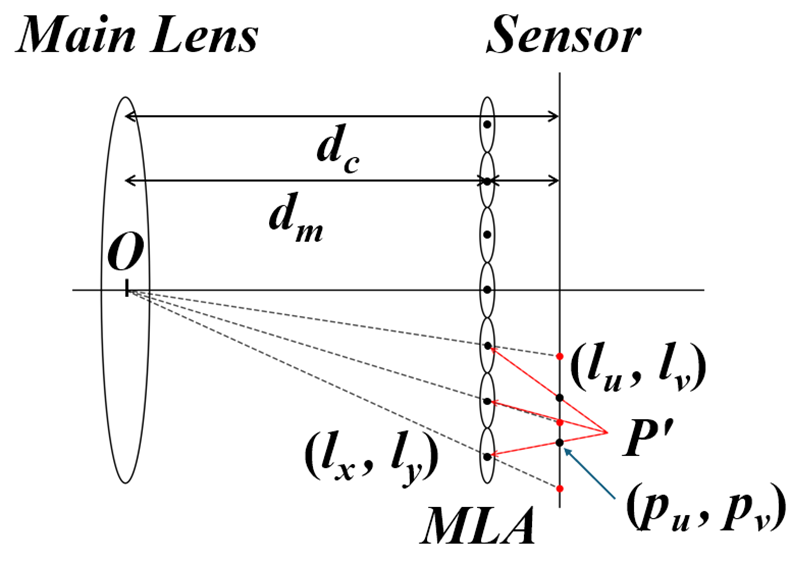}%
\label{fig:MIAproj}}
\caption{Decoupled projection model of light field camera. (a) for the main lens. (b) for the MLA.}
\label{fig_sim}
\end{figure}

The function of the MLA is not to form an image of the object directly, but to create a secondary image of the virtual point $\mathbf{P}'$ formed by the main lens, as shown in Fig.~\ref{fig:MIAproj}. For a given microlens with a known optical center, the projection of the virtual point through that microlens onto the sensor can be treated as pinhole imaging \cite{2013johannsen}. Let $d_m$ denote the distance from the MLA plane to the main lens, and $d_c$ the distance from the sensor plane to the main lens. For a microlens centered at $(l_x, l_y, d_m)^T$, its corresponding sensor image point is $(p_u, p_v, d_c)^T$. By the pinhole principle, the microlens center, the sensor image point, and the virtual point $\mathbf{P}' = (X'_c, Y'_c, Z'_c)^T$ are collinear. Since the MLA plane and the sensor plane are parallel, the Intercept Theorem gives:
\begin{equation}
\frac{l_x-p_u}{X'_c-l_x}=\frac{l_y-p_v}{Y'_c-l_y}=\frac{d_m-d_c}{Z'_c-d_m}
\label{eq:collinear}
\end{equation}

From Eq.~\ref{eq:collinear}, the sensor image coordinates can be expressed as a linear combination of the microlens center and the virtual point:
\begin{equation}
\begin{bmatrix}
p_u \\
p_v
\end{bmatrix}
=
\frac{Z'_c-d_c}{Z'_c-d_m}
\begin{bmatrix}
l_x \\
l_y
\end{bmatrix}
+
\frac{d_c -d_m}{Z'_c-d_m}
\begin{bmatrix}
X'_c \\
Y'_c
\end{bmatrix}
\label{eq:mla_intercept}
\end{equation}

Define the coefficient of the microlens center term as $\alpha$:
\begin{equation}
\alpha = \frac{Z'_c-d_c}{Z'_c-d_m}
\label{eq:definition}
\end{equation}

Equation~\ref{eq:mla_intercept} then simplifies to a compact convex-combination form:
\begin{equation}
\begin{bmatrix}
p_u \\
p_v
\end{bmatrix}
=
\alpha
\begin{bmatrix}
l_x \\
l_y
\end{bmatrix}
+
(1-\alpha)
\begin{bmatrix}
X'_c \\
Y'_c
\end{bmatrix}
\label{eq:mla_compact}
\end{equation}

We now prove that $\alpha$ is a linear fractional transformation (LFT) of $Z'_c$. A linear fractional transformation is defined as:
\begin{equation}
f(x) = \frac{ax + b}{cx + d}, \quad ad - bc \neq 0
\label{eq:lft_def}
\end{equation}
For $\alpha(Z'_c) = (Z'_c-d_c)/(Z'_c-d_m)$, we identify $a=1$, $b=-d_c$, $c=1$, $d=-d_m$, yielding $ad-bc = d_c-d_m$. Since the MLA and sensor planes are physically separated, $d_c \neq d_m$ holds by construction. If $d_c = d_m$, then $\alpha \equiv 1$ and Eq.~\ref{eq:mla_compact} would reduce to $(p_u, p_v)^T = (l_x, l_y)^T$, making the image position entirely independent of the virtual point, which contradicts the imaging principle of any MLA-based camera. Hence $ad-bc \neq 0$ is guaranteed, and $\alpha$ is indeed a linear fractional transformation of $Z'_c$. We therefore refer to the proposed formulation as the LFT model, and to $\alpha$ as the LFT coefficient.

The LFT coefficient also provides a direct criterion for identifying the optical configuration. For a Keplerian configuration, the MLA is placed behind the focal plane, closer to the sensor. Since objects are generally at a distance, the virtual point lies near the focal plane, giving $Z'_c > d_m > d_c$. Consequently:
\begin{equation}
\alpha = \frac{Z'_c-d_c}{Z'_c-d_m} > 1 \qquad \text{(Keplerian)}
\label{eq:keplerian}
\end{equation}
By analogous reasoning, a Galilean configuration yields $0 < \alpha < 1$. Thus, the estimated value of $\alpha$ serves as an auxiliary criterion for assessing the system configuration, whereas existing alternatives such as $K_1$ and $K_2$ \cite{2017Nousias} lack this direct physical interpretability.

Following the projection pipeline, the image coordinates are converted to pixel coordinates $(u,v)^T$ by multiplying the scale factors $s_x$ and $s_y$. Ignoring the principal point offset, Eq.~\ref{eq:mla_compact} becomes:
\begin{equation}
\begin{bmatrix}
u \\
v
\end{bmatrix}
=
\alpha
\begin{bmatrix}
m_x \\
m_y
\end{bmatrix}
+
(1-\alpha)
\begin{bmatrix}
q_x \\
q_y
\end{bmatrix}
\label{eq:LFT}
\end{equation}
where $(u,v)^T$ is the final image point and $(m_x, m_y)^T$, $(q_x, q_y)^T$ are the pixel-space quantities defined in Table~\ref{tab:notation}.

In Eq.~\ref{eq:LFT}, $(q_x, q_y)^T$ is common to all microlenses that observe the same object point, while $(m_x, m_y)^T$ varies per microlens. The virtual point is recovered analytically from clustered feature points, as detailed in Section~\ref{sec:analytical_solution}. For the microlens center, common practice \cite{2013Dansereau} models it via the microlens image of the main lens optical center (point $(l_u, l_v)^T$ in Fig.~\ref{fig_sim}), with the relation $(m_x, m_y)^T = d_m/d_c \cdot (l_u, l_v)^T$. This approach couples the microlens center estimation with other intrinsic parameters, leading to a complex calibration loop \cite{2020Liu}. To avoid this coupling, we estimate the microlens centers directly from the MLA geometry.

The following discussion assumes a square microlens arrangement; the hexagonal case is addressed in the appendix. Regardless of the pattern, there always exists a microlens whose optical center lies close to the main lens optical axis. For this microlens, the optical axis position on the sensor approximates the microlens center \cite{2000Zhang}, and we denote it as $(s_{\text{base}}, t_{\text{base}})^T$ in pixel coordinates. Let $r$ be the microlens diameter in pixels, obtainable from the MLA manufacturer specifications, and let $(\Delta i, \Delta j)$ be the index offset of element $(i,j)$ relative to the reference microlens. Neglecting rotation, the microlens center is:
\begin{equation}
\begin{bmatrix}
m_x \\
m_y
\end{bmatrix}
=
\begin{bmatrix}
s_{\text{base}} \\
t_{\text{base}}
\end{bmatrix}
+
r
\begin{bmatrix}
\Delta i \\
\Delta j
\end{bmatrix}
\end{equation}
Accounting for a slight in-plane rotation $\mathbf{R}(\theta)$ of the MLA due to assembly \cite{2013Dansereau}, the final microlens center in pixel coordinates is:
\begin{equation}
\begin{bmatrix}
m_x \\
m_y
\end{bmatrix}
=
\mathbf{R}(\theta)
\cdot
\begin{bmatrix}
s_{\text{base}} + r \cdot \Delta i \\
t_{\text{base}} + r \cdot \Delta j
\end{bmatrix}
\label{eq:grid_final}
\end{equation}

Once the microlens centers are established, the dimensions of each micro-image in the MIA can be determined from $H$ and $W$, the overall image height and width:
\begin{equation}
d_h = \frac{H}{n_h}, \quad d_v = \frac{W}{n_v}
\label{eq:mia_dim}
\end{equation}
where $n_h$ and $n_v$ are the numbers of microlenses along the corresponding directions. A white image \cite{2013Dansereau} is then used to refine these estimates: the coarse grid defined by Eqs.~\ref{eq:grid_final}--\ref{eq:mia_dim} is applied to segment the white image, and the circular boundary centroid within each segmented region replaces the initial microlens center, compensating for residual misalignment \cite{2022mathieu}. This refinement is carried out entirely during image preprocessing and does not involve any camera intrinsic parameters.

In contrast to the conventional workflow that first detects micro-image centers from the white image and then recovers microlens centers through the calibrated camera model \cite{2020Liu}, our approach decouples the two by estimating microlens centers from the MLA geometry alone. The microlens centers are thus available before the calibration of the main lens, eliminating a major source of parameter coupling.

In summary, Eqs.~\ref{eq:mla_compact}--\ref{eq:definition} reveal the central insight of the LFT model: for any single virtual point, the MLA imaging process reduces to a convex combination of a fixed microlens center and the virtual point itself, parameterized entirely by a single scalar $\alpha(Z'_c)$. The matrix form of this decoupled projection is presented in Section~\ref{sec:complete_model}.

\subsection{Projection Model of the Main Lens and Distortion}
\label{sec:mainlens_distortion}

When the MLA is omitted, the light field camera is equivalent to a standard camera \cite{liu2020stepwise}. Regardless of the plenoptic type, the main lens can always be described by the thin-lens model, as illustrated in Fig.~\ref{fig:Mainlensproj}. We define the camera coordinate system $\{ \mathbf{C}\}$ with its origin at the optical center of the main lens, the $Z$-axis pointing toward the scene along the optical axis, and the $X$ and $Y$ axes parallel to the sensor plane. For an object point $\mathbf{P} = (X_c, Y_c, Z_c)^T$ in $\{ \mathbf{C}\}$, its virtual image point is $\mathbf{P}' = (X'_c, Y'_c, Z'_c)^T$, given by the Gaussian lens formula:
\begin{equation}
\begin{bmatrix}
X'_c \\
Y'_c \\
Z'_c
\end{bmatrix}
=
\frac{F}{Z_c-F}
\begin{bmatrix}
X_c \\
Y_c \\
Z_c
\end{bmatrix}
\label{eq:Mainlens}
\end{equation}
where $F$ is the focal length of the main lens. It is worth noting that the main lens imaging process exhibits homography \cite{2000Zhang}: one object point maps to exactly one virtual point. In contrast, the MLA imaging process derived above is non-homographic --- one virtual point projects to multiple image points through different microlenses --- which is the mechanism by which light field cameras capture angular information.

Regarding distortion, we follow the common strategy in light field research by considering only the distortion of the main lens using the Brown-Conrady model \cite{2022mathieu,2000Zhang}. The MLA distortion is typically negligible \cite{2022mathieu} and is therefore omitted. A distorted virtual point $\mathbf{P}'_d = \phi(\mathbf{P}') = (p'_{dx}, p'_{dy})^T$ is computed from the distortion-free point $\mathbf{P}'=(X'_c, Y'_c)^T$ as:
\begin{equation}
\left\{
\begin{array}{ll}
p'_{dx} = X'_c \left( 1 + k_1 r^2 + k_2 r^4 \right) & \text{[radial]} \\
\quad + t_1 \left( r^2 + 2{X'_c}^2 \right) + 2 t_2 X'_c Y'_c & \text{[tangential]} \\
p'_{dy} = Y'_c \left( 1 + k_1 r^2 + k_2 r^4 \right) & \text{[radial]} \\
\quad + t_2 \left( r^2 + 2{Y'_c}^2 \right) + 2 t_1 X'_c Y'_c & \text{[tangential]}
\end{array}
\right.
\label{eq:dis}
\end{equation}
where $r^{2} = {X'_c}^{2} + {Y'_c}^{2}$. The radial distortion coefficients are $\{k_{1}, k_{2}\}$, and the tangential coefficients are $\{t_{1}, t_{2}\}$.

\subsection{Complete Projection Model}
\label{sec:complete_model}

Integrating the main lens, MLA, and distortion models with Zhang's calibration framework \cite{2000Zhang}, the complete projection from a 3D world point $\mathbf{P}_W$ to a 2D image point is expressed as:
\begin{equation}
\mathbf{H} =
\mathbf{H}_\alpha \cdot
\phi(\mathbf{K}
[\mathbf{R}|\mathbf{t}])
\label{eq:model}
\end{equation}
where $\mathbf{K}$ is the intrinsic matrix of the main lens, $[\mathbf{R}|\mathbf{t}]$ comprises the extrinsic rotation and translation, $\phi(\cdot)$ is the distortion function, and $\mathbf{H}_\alpha$ characterizes the MLA projection. According to Eq.~\ref{eq:LFT}, the matrix $\mathbf{H}_\alpha$ is:
\begin{equation}
\mathbf{H}_\alpha =
\begin{bmatrix}
(1-\alpha) & 0 & \alpha\, m_x \\
0 & (1-\alpha) & \alpha\, m_y \\
0 & 0 & 1
\end{bmatrix}
\label{eq:H_alpha}
\end{equation}

The construction of $\mathbf{H}_\alpha$ introduces only two supplementary parameters beyond the standard camera model: the microlens center $(m_x, m_y)^T$ and the LFT coefficient $\alpha$. Crucially, both can be estimated directly from raw image features via a closed-form least-squares solution (see Section~\ref{sec:analytical_solution}), without participating in the joint nonlinear optimization of the main lens parameters. Consequently, the entire projection model is decoupled, and the calibration of the main lens and the MLA can be carried out independently. Table~\ref{tab:lft_parameters} summarizes all parameters to be calibrated.

\begin{table}[htbp]
\centering
\caption{Parameters to be calibrated in the LFT model.}
\label{tab:lft_parameters}
\begin{tabular}{lc}
\toprule
\textbf{Parameter} & \textbf{DOF} \\
\midrule
Main lens intrinsic $\mathbf{K}$ & 5 \\
Main lens distortion $\phi(\cdot)$ & 4 \\
$d_c$ & 1 \\
$d_m$ & 1 \\
Extrinsic $[\mathbf{R}|\mathbf{t}]$ & $6 \times N_s$ \\
\bottomrule
\end{tabular}
\end{table}

\noindent where $N_s$ denotes the number of calibration scenes.

Nousias et al.\ \cite{2017Nousias} introduced two parameters $K_1$ and $K_2$ to describe the MLA. The equivalence with our formulation is:
\begin{equation}
K_1 = -\frac{(d_m+F)d_c}{(d_m-d_c)F}, \quad K_2 = \frac{d_m d_c}{d_m-d_c}
\label{eq:K1K2}
\end{equation}
However, $K_1$ and $K_2$ are highly coupled with the main lens focal length $F$. In their framework, $K_1$ and $K_2$ must be treated as special intrinsic parameters and estimated through joint optimization with all other parameters. This coupling increases the degrees of freedom in the solution space, leading to unstable calibration results. Moreover, obtaining physically meaningful parameters from $K_1$ and $K_2$ requires a secondary conversion step.

The LFT coefficient $\alpha$ differs from $K_1$ and $K_2$ in several fundamental respects. First, $\alpha$ is estimated directly from feature points via a closed-form least-squares solution (Section~\ref{sec:analytical_solution}), completely decoupled from the main lens intrinsics. Second, $\alpha$ possesses clear physical meaning: its value directly indicates the camera configuration (Galilean vs.\ Keplerian, see Eq.~\ref{eq:keplerian}), whereas $K_1$ and $K_2$ lack this interpretability.

\section{Light Field Calibration}

In this section, we present the calibration procedure based on the proposed LFT model. The method consists of four main steps: an analytical least-squares solution for the LFT coefficient and virtual point coordinates from clustered corner data, independent calibration of the main lens using the recovered virtual image points, fitting of the LFT parameters $d_c$ and $d_m$, and a final joint nonlinear refinement of all intrinsic parameters.

\subsection{Analytical Solution of $\mathbf{H}_\alpha$}
\label{sec:analytical_solution}

Following corner detection \cite{li2015automatic} and DBSCAN-based clustering \cite{2022mathieu} of the raw light field image, each resulting cluster $A_k$ contains $N_m$ image points $\{(u_i, v_i)^T\}$, $i=1,\dots,N_m$, that originate from the same checkerboard corner. By the homography of the main lens, each cluster $A_k$ corresponds to a single virtual point $(X'_c, Y'_c, Z'_c)^T$, whose $Z'_c$ coordinate is unknown but fixed for the cluster. Consequently, the LFT coefficient $\alpha$ is no longer a variable but an unknown constant $\alpha_k$ specific to that cluster.

For cluster $A_k$, the microlens center $(m_i^x, m_i^y)^T$ corresponding to each image point is known from Eq.~\ref{eq:grid_final}. Substituting into Eq.~\ref{eq:LFT}, we obtain $2N_m$ linear equations in three unknowns $(\alpha_k, q_x, q_y)$:
\begin{equation}
\begin{aligned}
u_i &= \alpha_k \, m_i^x + (1 - \alpha_k) \, q_x, \\
v_i &= \alpha_k \, m_i^y + (1 - \alpha_k) \, q_y, \quad i = 1, \dots, N_m.
\end{aligned}
\label{eq:cluster_equations}
\end{equation}

Although Eq.~\ref{eq:cluster_equations} is bilinear in $(\alpha_k, q_x, q_y)$, it becomes linear in the auxiliary variables $(\alpha_k,\; (1-\alpha_k)q_x,\; (1-\alpha_k)q_y)$. A solution exists when $2N_m \geq 3$, requiring at least two image points per cluster. Solving the resulting $2N_m \times 3$ least-squares system and converting back yields the closed-form estimates:
\begin{equation}
\hat{\alpha}_k = \frac
{N_m \sum (m_i^x u_i + m_i^y v_i) - \sum m_i^x \sum u_i - \sum m_i^y \sum v_i}
{N_m \sum \big((m_i^x)^2 + (m_i^y)^2\big) - \big(\sum m_i^x\big)^2 - \big(\sum m_i^y\big)^2}
\label{eq:alpha_closed}
\end{equation}
\begin{equation}
\hat{q}_x = \frac{\sum u_i - \hat{\alpha}_k \sum m_i^x}{N_m (1 - \hat{\alpha}_k)},
\quad
\hat{q}_y = \frac{\sum v_i - \hat{\alpha}_k \sum m_i^y}{N_m (1 - \hat{\alpha}_k)}.
\label{eq:virtual_closed}
\end{equation}

Applying Eqs.~\ref{eq:alpha_closed}--\ref{eq:virtual_closed} to every cluster establishes the mapping $A_k \leftrightarrow (\hat{\alpha}_k, \hat{q}_x, \hat{q}_y)^T$. The recovered virtual image points form a virtual sensor image of the checkerboard, which can be automatically sorted using the energy-based method of \cite{li2015automatic}. This establishes the complete mapping from each checkerboard corner $[x, y, 0]^T$ to its observed image points via the corresponding virtual point. The matrix $\mathbf{H}_\alpha$ for each microlens is then fully determined by substituting $\hat{\alpha}_k$ and the microlens center from Eq.~\ref{eq:grid_final} into Eq.~\ref{eq:H_alpha}.

\subsection{Main Lens Calibration}

With the virtual image points $(\hat{q}_x, \hat{q}_y)^T$ serving as sensor observations, the light field camera reduces to an equivalent standard camera. We place the world coordinate frame on the checkerboard plane ($z=0$) \cite{2000Zhang} and estimate the main lens intrinsic matrix $\mathbf{K}$ and the per-view extrinsics $[\mathbf{R}|\mathbf{t}]$ by minimizing the reprojection error on the virtual image:
\begin{equation}
\underset{\mathbf{K}, [\mathbf{R}|\mathbf{t}]}{\arg\min} \sum \Big\| (\hat{q}_x, \hat{q}_y)^T - \phi\big(\mathbf{K}[\mathbf{R}|\mathbf{t}]\, [x, y, 0]^T\big) \Big\|^2.
\label{eq:mainlens_opt}
\end{equation}

The optimization is performed using the Levenberg-Marquardt (LM) algorithm. The intrinsic matrix $\mathbf{K}$ is initialized with the theoretical focal length $f$ and the image center $(c_x, c_y)^T$, while distortion coefficients are initialized to zero, following the standard procedure of Zhang \cite{2000Zhang}. Notably, Eq.~\ref{eq:mainlens_opt} is entirely independent of the LFT coefficient $\alpha$, which embodies the key decoupling property of the proposed model: the main lens can be calibrated without any knowledge of the MLA parameters beyond the virtual point estimates obtained analytically from the clusters.

\subsection{Estimation of $d_c$ and $d_m$}

Once the per-view extrinsics are known, the depth $Z'_c$ of each virtual point is obtained from the extrinsics and Eq.~\ref{eq:Mainlens}. For a checkerboard corner at $[x, y, 0]^T$ under extrinsics $[\mathbf{R}|\mathbf{t}]$, the $Z$-coordinate in the camera frame is
\begin{equation}
Z'_c = \frac{F(r_{31} x + r_{32} y + t_z)}{r_{31} x + r_{32} y + t_z - F},
\label{eq:zc_from_extrinsic}
\end{equation}
where $r_{ij}$ and $t_z$ denote the elements of $\mathbf{R}$ and $\mathbf{t}$, respectively.

Each cluster provides an observed pair $(\hat{\alpha}_k, Z'_c)$. From the definition $\alpha = (Z'_c - d_c)/(Z'_c - d_m)$ (Eq.~\ref{eq:definition}), rearranging yields the linear relation $-d_c + \alpha d_m = Z'_c(\alpha - 1)$. Stacking the $M$ observations across all corners and views, $d_c$ and $d_m$ are estimated by least squares:
\begin{equation}
\begin{bmatrix}
\hat{d}_c \\
\hat{d}_m
\end{bmatrix} = (\mathbf{A}^T \mathbf{A})^{-1} \mathbf{A}^T \mathbf{z},
\label{eq:estifordcdm}
\end{equation}
where
\[
\mathbf{A} = \begin{bmatrix}
-1 & \hat{\alpha}_1 \\
-1 & \hat{\alpha}_2 \\
\vdots & \vdots \\
-1 & \hat{\alpha}_M
\end{bmatrix},
\quad
\mathbf{z} = \begin{bmatrix}
Z'_{c,1}(\hat{\alpha}_1 - 1) \\
Z'_{c,2}(\hat{\alpha}_2 - 1) \\
\vdots \\
Z'_{c,M}(\hat{\alpha}_M - 1)
\end{bmatrix}.
\]
where $M$ denotes the total number of corner observations.

\subsection{Joint Nonlinear Refinement}

With all parameters initialized, a final joint optimization minimizes the complete light field reprojection error:
\begin{equation}
g(\mathbf{R}, \mathbf{t}, \mathbf{K}, \phi(\cdot), \mathbf{H}_\alpha) = \sum \| \mathbf{p} - \mathbf{p}_r \|^2,
\label{eq:joint_cost}
\end{equation}
where $\mathbf{p}$ is a detected image point and $\mathbf{p}_r = \mathbf{H} \mathbf{p}$ is its projection through the full model (Eq.~\ref{eq:model}). The MLA center positions defined by Eq.~\ref{eq:grid_final} remain constant throughout this refinement, further reducing the optimization degrees of freedom. The LM algorithm is used to minimize the cost function.

The calibration workflow is summarized as follows. First, corner detection and clustering produce the sets $A_k$. Second, $\hat{\alpha}_k$ and the virtual image points are computed analytically via Eqs.~\ref{eq:alpha_closed}--\ref{eq:virtual_closed}. Third, the main lens intrinsics and extrinsics are calibrated on the virtual image (Eq.~\ref{eq:mainlens_opt}), independently of $\alpha$. Fourth, $d_c$ and $d_m$ are fitted from the $(\hat{\alpha}_k, Z'_c)$ pairs (Eq.~\ref{eq:estifordcdm}). Finally, all parameters are jointly refined by minimizing Eq.~\ref{eq:joint_cost}. This stepwise procedure, enabled by the decoupled structure of the LFT model, avoids the strong parameter coupling that complicates prior calibration methods.

\section{Experimental results}

\subsection{Datasets and Evaluation Metrics}

We adopt the publicly available R12-series datasets provided by Labussi\`ere et al.\ \cite{2022mathieu}, which consist of three configurations with different focus distances $h$: $R12$-$A$ ($h=450$\,mm), $R12$-$B$ ($h=1000$\,mm), and $R12$-$C$ ($h=\infty$). These datasets were selected because they are recent, widely validated, and captured with a physical light field camera. Each configuration contains approximately $30$ scenes for free-hand calibration and a set of images acquired under controlled translation along the $Z$-axis for quantitative evaluation, with ground-truth displacements measured by a Leica ScanStation P20 laser scanner.

The acquisition hardware is a Raytrix R12 focused light field camera with a multi-focus hexagonal MLA. Its key specifications are listed in Table~\ref{tab:r12_specs}. Details of the calibration targets and the distance ranges used in each configuration are summarized in Table~\ref{tab:r12_parameters}.

\begin{table}[htbp]
\centering
\caption{Key specifications of the Raytrix R12 camera.}
\label{tab:r12_specs}
\begin{tabular}{ll}
\toprule
\textbf{Parameter} & \textbf{Value} \\
\midrule
MLA arrangement & Hexagonal \\
Number of microlens types & 3 \\
MLA dimensions & $176\;(\text{H}) \times 152\;(\text{V})$ \\
Main lens focal length & 50\,mm \\
Pixel size & $5.5\,\mu\text{m} \times 5.5\,\mu\text{m}$ \\
Sensor resolution & $4080\;(\text{H}) \times 3068\;(\text{V})$ \\
\bottomrule
\end{tabular}
\end{table}

\begin{table}[htbp]
\caption{Dataset parameters, calibration distances, and evaluation ranges.}
\label{tab:r12_parameters}
\centering
\begin{tabular}{ccccccc}
\toprule
 & $h$ & \textbf{Target} & \textbf{Scale} & \multicolumn{2}{c}{\textbf{Calibration}} & \textbf{Evaluation} \\
\cmidrule(lr){5-6}
 & (mm) & & (mm) & min & max & range (step) \\
\midrule
$R12$-$A$ & 450 & $9 \times 5$ & 10  & 175 & 400 & 265--385 (10)  \\
$R12$-$B$ & 1000 & $8 \times 5$ & 20  & 400 & 775 & 450--900 (50)  \\
$R12$-$C$ & $\infty$ & $6 \times 4$ & 30  & 500 & 2500 & 400--1250 (50)  \\
\bottomrule
\end{tabular}
\end{table}

\noindent We evaluate calibration accuracy using three complementary metrics. The reprojection error is the root mean square (RMS) distance between the detected corner points and their projections through the calibrated model:
\begin{equation}
\text{RMS} = \sqrt{\frac{1}{N}\sum_{i=1}^{N} \|\mathbf{p}_i - \hat{\mathbf{p}}_i\|^2},
\label{eq:rms}
\end{equation}
where $\mathbf{p}_i$ denotes a detected image point and $\hat{\mathbf{p}}_i$ its predicted projection. For the controlled-translation sequences, the \textbf{relative translation error} $\epsilon_z$ quantifies the accuracy of the extrinsics:
\begin{equation}
\epsilon_z = \frac{|\hat{\delta z} - \delta z|}{\delta z} \times 100\%,
\label{eq:relatederr}
\end{equation}
where $\delta z$ is the ground-truth displacement and $\hat{\delta z}$ the estimated displacement, computed following the protocol of \cite{2022mathieu}. The \textbf{coefficient of determination} $R^2$ measures the goodness of fit of the LFT relationship between $\alpha$ and $Z'_c$:
\begin{equation}
R^2 = 1 - \frac{SS_{\text{res}}}{SS_{\text{tot}}},
\label{eq:r2}
\end{equation}
where $SS_{\text{res}}$ and $SS_{\text{tot}}$ are the residual and total sum of squares, respectively.

\subsection{Physical Camera}

We first verify the core prediction of the LFT model --- the fractional linear relationship between $\alpha$ and $Z'_c$ --- then evaluate calibration accuracy from two complementary perspectives: the image-space reprojection error (Section~\ref{sec:calibration}) and the object-space translation estimation error (Section~\ref{sec:quantitative}).

\begin{figure}[ht]
    \centering
    \subfloat[$R12$-$A$]{\includegraphics[width=0.45\linewidth]{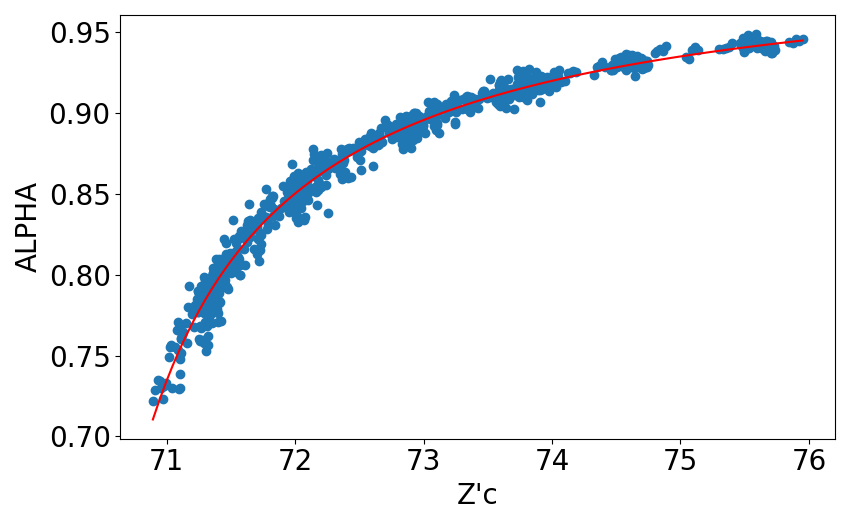}%
    }
    \hfil
    \subfloat[$R12$-$B$]{\includegraphics[width=0.45\linewidth]{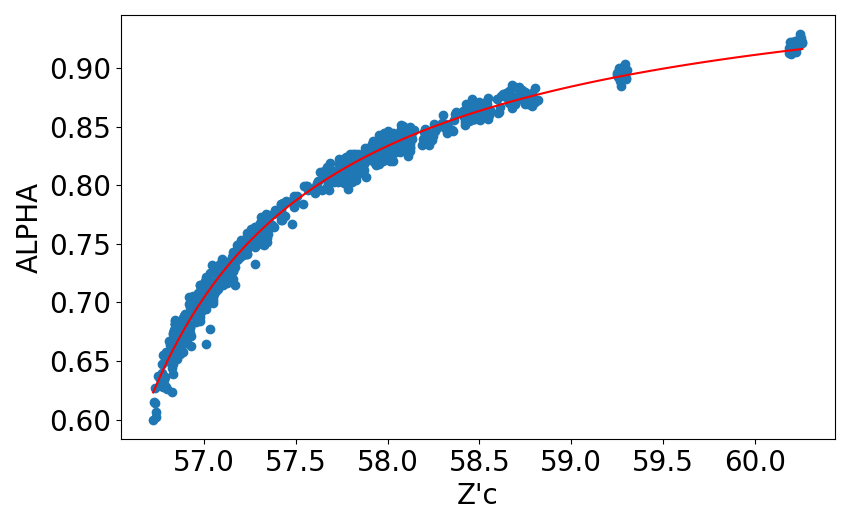}%
    }

    \subfloat[$R12$-$C$]{\includegraphics[width=0.45\linewidth]{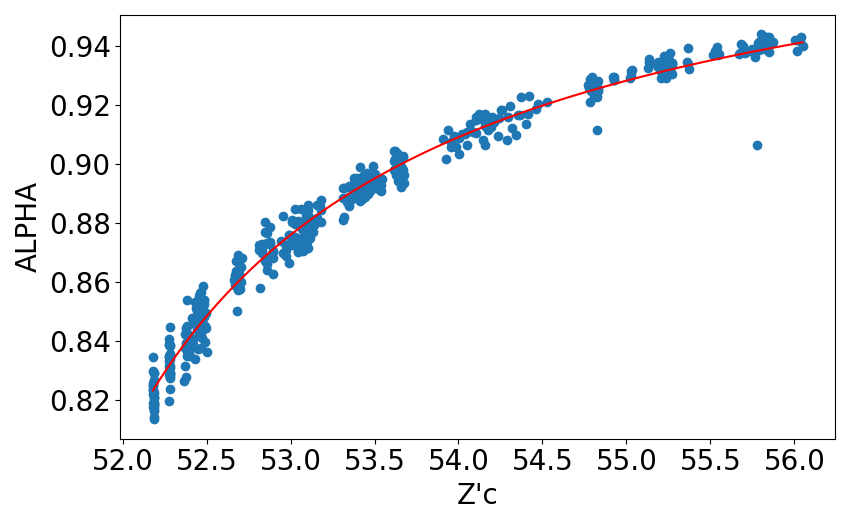}%
    }

    \caption{Variation of the parameter $\alpha$ with $Z'_c$ for different datasets.}
    \label{fig:alphawithzv}
\end{figure}

To verify the $\alpha$--$Z'_c$ relationship, $Z'_c$ is computed from the extrinsics obtained via $PnP$ and Eq.~\ref{eq:Mainlens}, while $\hat{\alpha}$ is estimated from the corner clusters using Eqs.~\ref{eq:alpha_closed}--\ref{eq:virtual_closed}. The results for all three datasets are shown in Fig.~\ref{fig:alphawithzv}. The red curves represent the theoretical LFT functions using the estimated $\hat{d}_c$ and $\hat{d}_m$ from Eq.~\ref{eq:estifordcdm}. The $R^2$ values for the three datasets are $0.9848$ ($R12$-$A$), $0.9907$ ($R12$-$B$), and $0.9846$ ($R12$-$C$), indicating excellent agreement between the model predictions and the observed data. These results confirm that the LFT relationship defined in Eq.~\ref{eq:definition} indeed holds in physical light field data.

Furthermore, the statistical analysis reveals that all estimated $\alpha$ values across the three datasets are less than $1$. According to the optical configuration criterion derived in Eq.~\ref{eq:keplerian}, $\alpha<1$ implies a Galilean configuration. This is consistent with the documented design of the Raytrix R12 camera, which is indeed of Galilean type, thereby corroborating the physical validity of the LFT coefficient definition.

\subsubsection{Free-Hand Camera Calibration}
\label{sec:calibration}

Of the approximately $30$ scenes per dataset, $20$ images satisfied the requirement that each object point corresponds to at least two image points. The list of images used is available in the open-source code.

Since our model calibrates the light field camera in the raw image, we compare our intrinsics with the SOTA result reported in \cite{2022mathieu,2017Noury} and the method of Nousias et al.\cite{2017Nousias} which provides a set of intrinsics for each micro lens type. The equivalence of our parameters and their parameters is given by:
\begin{equation}
\begin{aligned}
F &= \frac{k_{11}s_x + k_{22}s_y}{2}, \quad u_0 = k_{13}, \quad v_0 = k_{23} \\
D &= d_c, \quad d = d_c - d_m
\end{aligned}
\end{equation}
where $k_{ij}$ are elements of the intrinsic matrix $\mathbf{K}$ of main lens. 

\begin{table}[h]
\centering
\small
\caption{Comparison of methods for $R12-A$ ($F = 50mm$, $h = 450mm$)}
\begin{tabular}{l c c c c c c}
\toprule
 & Ours. & \cite{2022mathieu} & \cite{2017Noury} & \multicolumn{3}{c}{\cite{2017Nousias}} \\
\cmidrule(lr){5-7}
 &  &  &  & Type 1 & Type 2 & Type 3 \\
\midrule
$F$ & 61.010 & 49.714 & 54.888 & 61.305 & 62.476 & 63.328 \\
$D$ & 70.080 & 56.701 & 62.425 & 71.131 & 72.541 & 73.530 \\
$d$ & 0.338 & 0.324 & 0.402 & -- & -- & -- \\
$f^{(1)}$ & -- & 578.18 & -- & -- & -- & -- \\
$f^{(2)}$ & -- & 505.42 & -- & -- & -- & -- \\
$f^{(3)}$ & -- & 552.08 & -- & -- & -- & -- \\
$u_0$ & 2039.3 & 2070.9 & 2289.8 & 1984.9 & 2034.5 & 1973.7 \\
$v_0$ & 1898.2 & 1610.9 & 1528.2 & 1482.1 & 1481.0 & 1495.2 \\
\bottomrule
\end{tabular}
\label{tab:comparison_r12a}
\end{table}

\begin{table}[h]
\centering
\small
\caption{Comparison of methods for $R12-B$ ($F = 50mm$, $h = 1000mm$)}
\begin{tabular}{l c c c c c c}
\toprule
 & Ours. & \cite{2022mathieu} & \cite{2017Noury} & \multicolumn{3}{c}{\cite{2017Nousias}} \\
\cmidrule(lr){5-7}
 &  &  &  & Type 1 & Type 2 & Type 3 \\
\midrule
$F$ & 53.604 & 50.047 & 51.262 & 53.913 & 52.988 & 52.977 \\
$D$ & 56.142 & 52.125 & 53.296 & 56.062 & 55.128 & 55.124 \\
$d$ & 0.367 & 0.336 & 0.363 & -- & -- & -- \\
$f^{(1)}$ & -- & 580.49 & -- & -- & -- & -- \\
$f^{(2)}$ & -- & 504.31 & -- & -- & -- & -- \\
$f^{(3)}$ & -- & 546.36 & -- & -- & -- & -- \\
$u_0$ & 1758.79 & 1958.3 & 1934.9 & 2074.7 & 2094.7 & 1837.0 \\
$v_0$ & 1951.42 & 1802.9 & 1759.3 & 1640.2 & 1649.1 & 1620.4 \\
\bottomrule
\end{tabular}
\label{tab:comparison_r12b}
\end{table}

\begin{table}[h]
\centering
\small
\caption{Comparison of methods for $R12-C$ ($F = 50mm$, $h = \infty$)}
\begin{tabular}{l c c c c c c}
\toprule
 & Ours. & \cite{2022mathieu} & \cite{2017Noury} & \multicolumn{3}{c}{\cite{2017Nousias}} \\
\cmidrule(lr){5-7}
 &  &  &  & Type 1 & Type 2 & Type 3 \\
\midrule
$F$ & 51.032 & 50.013 & 53.322 & 51.113 & 49.919 & 50.812 \\
$D$ & 50.671 & 49.362 & 52.379 & 50.331 & 49.067 & 49.882 \\
$d$ & 0.333 & 0.307 & 0.319 & -- & -- & -- \\
$f^{(1)}$ & -- & 569.88 & -- & -- & -- & -- \\
$f^{(2)}$ & -- & 491.71 & -- & -- & -- & -- \\
$f^{(3)}$ & -- & 535.28 & -- & -- & -- & -- \\
$u_0$ & 1635.2 & 1692.1 & 2131.6 & 1966.3 & 1913.8 & 2052.5 \\
$v_0$ & 1556.4 & 1677.8 & 1445.9 & 1484.6 & 1487.2 & 1492.7 \\
\bottomrule
\end{tabular}
\label{tab:comparison_r12c}
\end{table}

Tables~\ref{tab:comparison_r12a}, \ref{tab:comparison_r12b} and \ref{tab:comparison_r12c} present the intrinsics of our method and those of the competing methods. $F$, $D$ and $d$ are in $mm$; $f^{(1)}$, $f^{(2)}$, and $f^{(3)}$ are in $\mu m$; $u_0$ and $v_0$ are in pixels. The intrinsics of~\cite{2022mathieu,2017Noury,2017Nousias} are taken directly from the corresponding papers.

Several observations can be drawn from the intrinsic calibration results. First, \cite{2022mathieu} yields the highest parameter consistency across the three datasets, particularly for $F$ and $D$ which remain nearly constant regardless of the focusing distance, confirming the findings reported in their paper. Second, the close alignment of our estimated $F$ and $D$ with those of \cite{2017Nousias} indirectly validates the correctness of the proposed model despite its simplified treatment of the MLA. Third, \cite{2017Nousias} requires separate intrinsic parameter sets for each of the three microlens types, resulting in the largest total number of parameters among all methods. Moreover, the principal point of \cite{2017Nousias} fluctuates even within the same dataset, a consequence of the strong coupling between their auxiliary parameters $K_1$, $K_2$ and the main lens intrinsics. In contrast, our method employs a single, unified set of intrinsic parameters with fewer degrees of freedom. The discrepancy in $(u_0,v_0)^T$ between our method and \cite{2017Nousias} reflects a fundamental structural difference: in the LFT model, the principal point is solely associated with the main lens, as the MLA projection has been fully decoupled.

The distortion parameters are reported in Table~\ref{tab:distcalib}. They are relatively large compared to the values reported in \cite{2022mathieu}. The primary reason is that the distortion in our model is associated only with the virtual image point estimated by the least-squares algorithm, causing the estimation error to accumulate in the distortion coefficients. The simulation-based evaluation in Section~\ref{sec:LFTsim} confirms this interpretation.
\begin{table}[htbp]
\centering
\caption{Distortions of $R12-A$, $R12-B$ and $R12-C$}
\label{tab:distcalib}
\begin{tabular}{l c c c}
\toprule
 & $R12-A$. & $R12-B$ & $R12-C$ \\
\midrule
$k_1$ & 2.92$\times$10$^{-2}$ & -1.69$\times$10$^{-3}$ & 5.39$\times$10$^{-2}$  \\
$k_2$ & -3.89$\times$10$^{-1}$ & -8.45$\times$10$^{-1}$ & -3.89$\times$10$^{-2}$ \\
$t_1$ & 1.10$\times$10$^{-2}$ & 1.16$\times$10$^{-2}$ & 9.29$\times$10$^{-3}$ \\
$t_2$ & 1.02$\times$10$^{-2}$ & -8.53$\times$10$^{-3}$ & -1.81$\times$10$^{-2}$ \\

\bottomrule
\end{tabular}
\end{table}

\subsubsection{Quantitative Evaluations of the Camera Model}
\label{sec:quantitative}

Having verified the LFT relationship and the intrinsic calibration quality, we now assess the model's accuracy from the object-space perspective using the controlled-translation sequences. We first examine the reprojection error of the main lens and the complete light field camera separately, 

A key finding is that the main lens model alone exhibits high reprojection error, with RMS values of $3.180$, $1.287$, and $2.367$ pixels for \mbox{$R12$-$A$}, \mbox{$R12$-$B$}, and \mbox{$R12$-$C$}, respectively. However, after applying the micro lens projection via $\mathbf{H_\alpha}$, all errors are reduced to below $1$ pixel ($0.716$, $0.833$, and $0.675$ pixels), meeting the fundamental accuracy requirement for light field camera calibration. Our method achieves the best reprojection accuracy on $R12$-$A$ and $R12$-$C$ among all compared methods, and ranks second on $R12$-$B$ (where \cite{2022mathieu} performs best). The method of \cite{2017Nousias} yields the lowest reprojection errors overall, but exhibits notable inconsistency across its three microlens types, with the reprojection error varying substantially between types. In contrast, our unified model delivers stable performance across all configurations. This result indicates two key points. Firstly, since our nonlinear optimization minimizes the final light field reprojection error, an increase in intermediate errors is acceptable. Secondly, the proposed $\mathbf{H_\alpha}$ matrix effectively characterizes the MLA imaging process. Importantly, it can even compensate for miscalibrations in the main lens model. We attribute the primary error source in the main lens model to our simplification of treating all micro lens types as identical.

\begin{figure*}[h]
\centering
\includegraphics[width=0.85\linewidth]{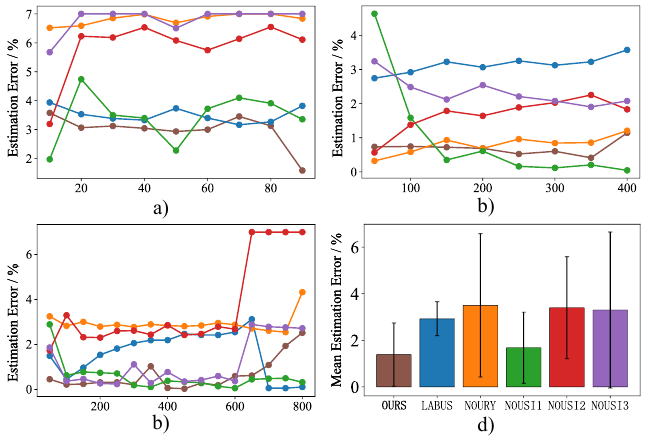}
\caption{Translation errors w.r.t the ground truth displacement for $R12-A$, $R12-B$ and $R12-C$.}
\label{fig:eval}
\end{figure*}

The translation errors along the $z$-axis with respect to the ground truth displacement, derived from the external parameter estimation, and the line graphs for each dataset are shown in Fig.~\ref{fig:eval}(a--c). The mean and standard deviation of $\epsilon_z$ across all datasets for each method are summarized in Fig.~\ref{fig:eval}(d) (color version recommended for visualization).

The per-method mean translation errors across all three datasets are as follows. Our method achieves $\epsilon_z = 1.38 \pm 1.35\%$, outperforming \cite{2022mathieu} at $\epsilon_z = 2.92 \pm 0.73\%$ and \cite{2017Noury} at $\epsilon_z = 3.50 \pm 3.08\%$. The best numerical result is obtained by \cite{2017Nousias} type~$1$, with $\epsilon_z = 1.68 \pm 1.53\%$. However, its exceptionally high standard deviation reveals substantial inconsistency across its different microlens types: type~$3$ of the same method reaches $\epsilon_z = 3.30 \pm 3.35\%$, making it the worst-performing configuration overall. Although the intrinsic parameters of our model are similar to those of \cite{2017Nousias} type~$1$, our method significantly outperforms it in terms of both mean accuracy and consistency in the quantitative evaluation.

We now analyze the per-dataset behavior based on the line graphs in Fig.~\ref{fig:eval}. All methods perform better on $R12$-$B$, likely because the working distance of this configuration is the most suitable for the current focal length setting. On this dataset, our method achieves the best performance, with the relative error consistently below $1.5\%$ and with small variation. On $R12$-$A$, \cite{2022mathieu} exhibits better stability than our method, while \cite{2017Noury} remains stable but incurs a larger overall error. On $R12$-$C$, our method shows a relatively large error at a displacement of $800$\,mm, whereas at other distances the error is smaller and comparable to that of \cite{2017Noury}. \cite{2022mathieu} also displays some fluctuation on this dataset. Overall, \cite{2022mathieu} rarely produces significant outliers, but its average error is higher than ours.

The instability of \cite{2017Nousias} can be attributed to the strong coupling between its auxiliary parameters $K_1$, $K_2$ and the main lens intrinsics. Since $K_1$ and $K_2$ must be jointly optimized with all other parameters, the solution space has higher degrees of freedom, leading to unstable calibration across different microlens types. This conclusion is consistent with the observed external parameter estimation errors. In contrast, the LFT coefficient in our model is estimated independently from the main lens parameters, eliminating this coupling and yielding more stable results. The method of \cite{2022mathieu} achieves better stability than ours, but at the cost of a higher mean error, which may be attributed to their complex model of light field blur features causing rounding-error accumulation. The method of \cite{2017Noury}, on the other hand, adopts an overly simplified model, resulting in both lower accuracy and reduced stability.

A limitation of our method is its dependence on each corner being observed by at least two microlenses to form a valid cluster. When this condition is not met, the overdetermined system in Eq.~\ref{eq:cluster_equations} becomes underdetermined and has no solution. This issue manifested in one frame each of the $R12$-$A$ and $R12$-$B$ datasets, where corner points located at the periphery of the checkerboard were captured by only a single microlens. These frames were excluded from calibration, yet the calibration still completed successfully with the remaining images. Future work may explore robust feature construction strategies that can incorporate isolated corner points, thereby strengthening the practical resilience of the calibration pipeline.

Overall, by decoupling the main lens and MLA projections, the proposed LFT model achieves a mean translation error of $1.38\%$, lower than all competing methods that use a unified parameter set, while maintaining sub-pixel reprojection accuracy.

\subsection{Simulated Data}

To acquire the ground truth for calibration, we generated synthetic scenes using the proposed LFT model. The simulation configuration is as follows: $F = 50$ $mm$, $d_c = 58$ $mm$, $d_m = 57$ $mm$. The target is a checkerboard with $9 \times 6$ of size $52.5$ $mm$, and the raw image size is $6500 \times 4700$ with the pixel size of $3.6$ $\mu m$. We generated $20$ scenes of random poses for calibration and $20$ with a controlled translation motion along the $Z$-axis for evaluation. All data are generated without distortion.

\subsubsection{Quantitative Evaluations of LFT model}
\label{sec:LFTsim}

\begin{figure}[h]
    \centering
    \includegraphics[width=0.80\linewidth]{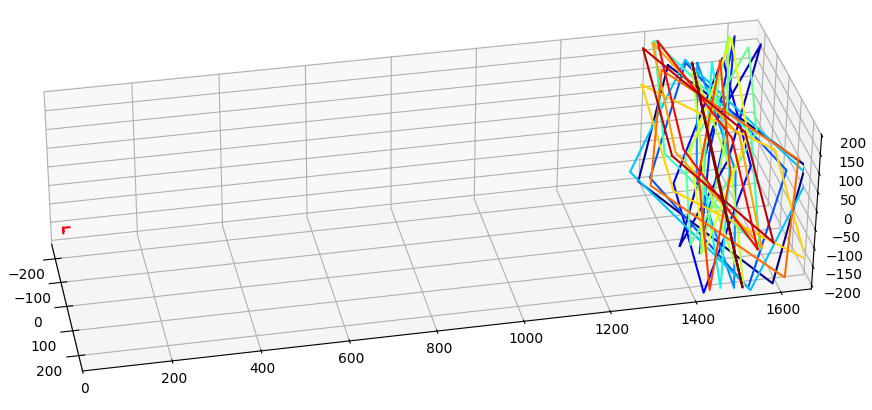}%
    \caption{Calibration result using simulated data.}
    \label{fig:simrt}
\end{figure}

Visualization of free-hand calibration is shown in Fig.~\ref{fig:simrt}, with the estimated intrinsic parameters shown in Table~\ref{tab:sim_eval}. The results show that the intrinsic parameters estimated by our algorithm are very close to the ground truth, with a mean relative error of $0.18\%$ and a standard deviation of $0.014$. The re-projection errors are also very small, with RMSE of $0.456$ pixels for the main lens and $0.418$ for the light field camera. The estimated distortion coefficients are $k_1 = -4.70 \times 10^{-2}$, $k_2 = 1.015$, $t_1 = -5.42\times 10^{-5} $, and $t_2 = 2.62 \times 10^{-4}$. Overall, the calibration demonstrates excellent accuracy and reliability.
\begin{table}[h]
\centering
\caption{The intrinsic parameters estimation results of the simulated data.}
\begin{tabular}{l c c c c c}
\toprule
Method & $F$ & $d_c$ & $d_m$ & $u_0$ & $v_0$ \\
\midrule
Ours & 50.032 & 57.913 & 56.929 & 3263.3 & 2354.3 \\
G.T. & 50.000 & 58.000 & 57.000 & 3250.0 & 2350.0 \\
\bottomrule
\end{tabular}
\label{tab:sim_eval}
\end{table}

\begin{figure*}[htbp]
    \subfloat[]{\includegraphics[width=0.32\linewidth]{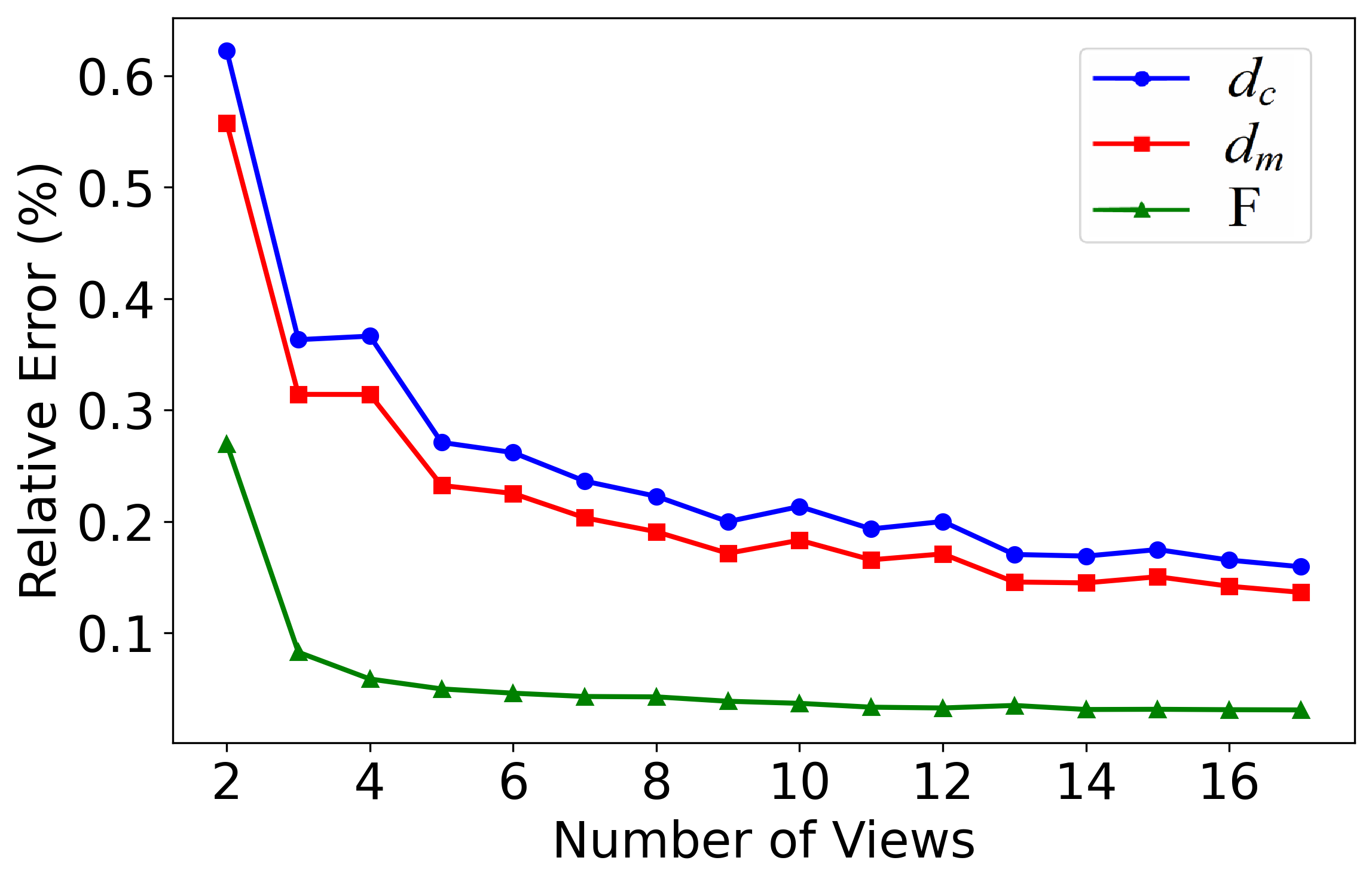}%
    \label{subfig:numofviews}
    }
    \hfil
    \subfloat[]{\includegraphics[width=0.31\linewidth]{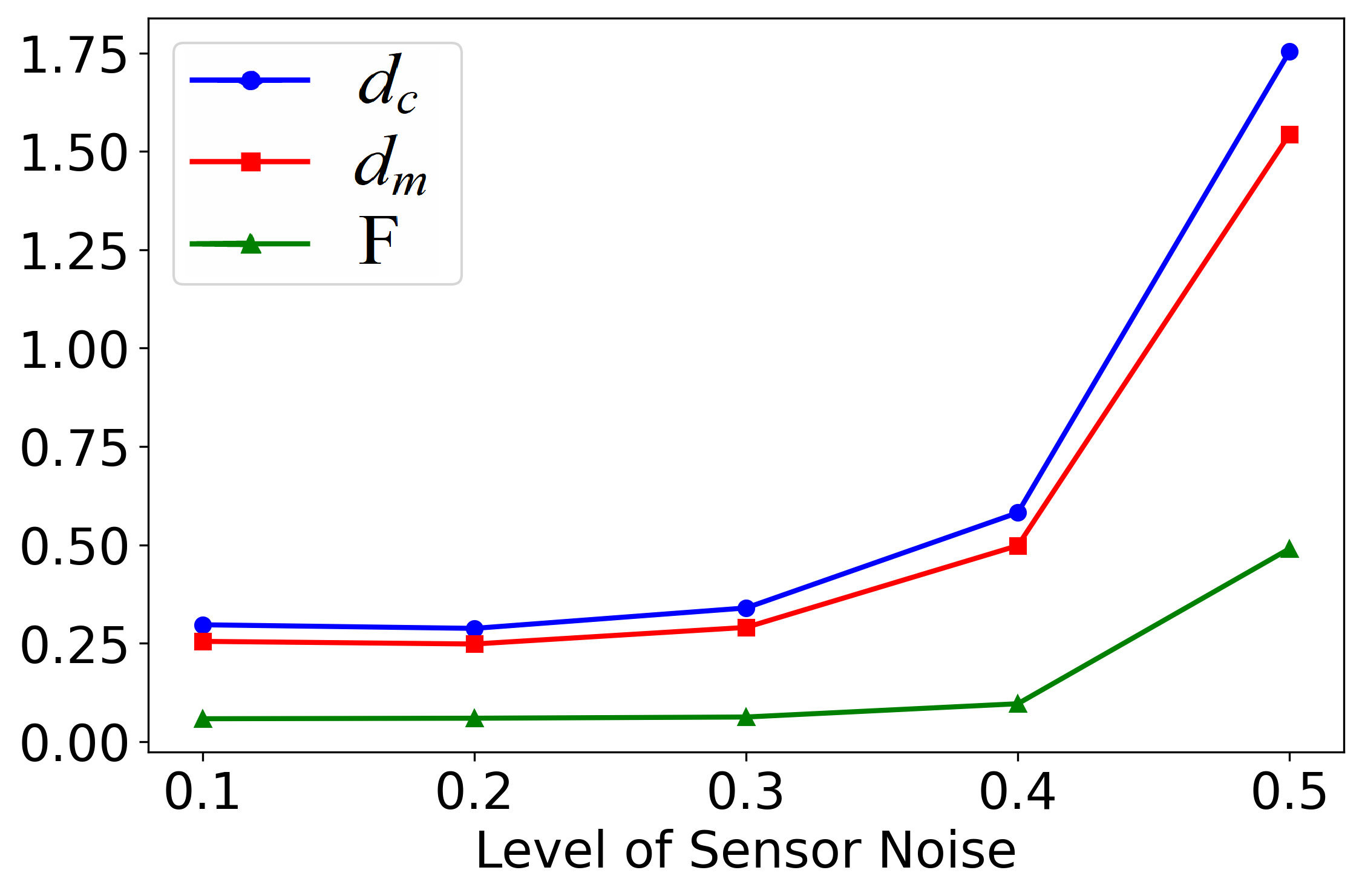}%
    \label{subfig:sensornoise}
    }
    \subfloat[]{\includegraphics[width=0.32\linewidth]{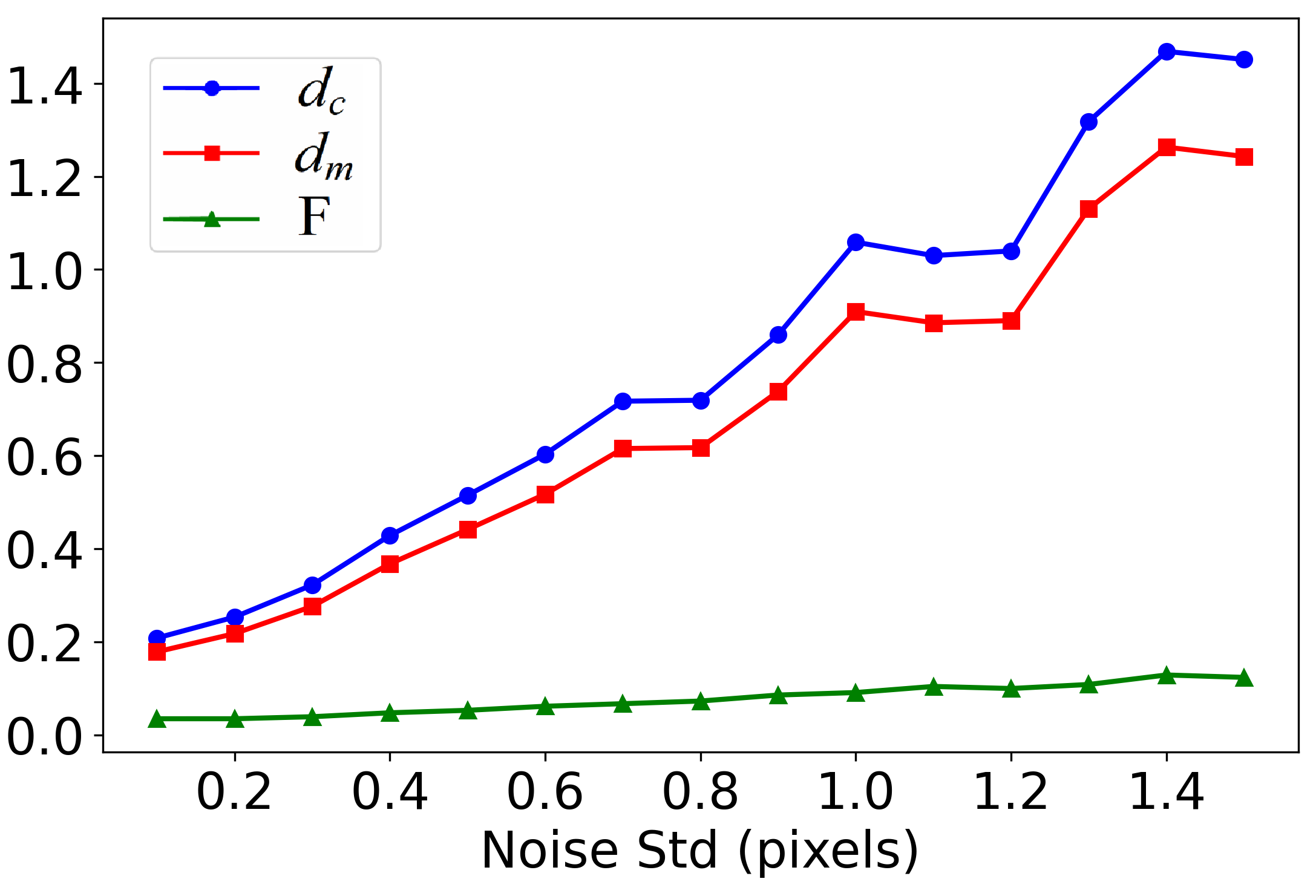}%
    \label{subfig:cripnoise}
    }
    \caption{(a) Relative error w.r.t. number of view. (b) Relative error w.r.t. sensor noise. (c) Relative error w.r.t. observation noise.}
    \label{fig:roubutnes}
\end{figure*}

\subsubsection{Calibration robustness analysis}

To analyze the robustness of the proposed algorithm, we first examined how calibration accuracy depends on the number of views, and then evaluated sensitivity to two types of noise. All ground truth data were generated using the configuration described in Section~\ref{sec:LFTsim}. The calibration target is a checkerboard with $9 \times 6$ corners and a square size of $52.5$\,mm. The sensor resolution is $6500 \times 4700$ pixels with a pixel size of $3.6$\,\textmu m. The target-to-camera distance ranges from $1350$\,mm to $1550$\,mm. Ray-tracing-based simulation \cite{2022mathieu} is employed to generate realistic light field images while excluding confounding variables.

\textbf{Dependence on the number of views.} We varied the number of calibration views from $2$ to $16$. For each count, $200$ independent trials were conducted, each with random plane poses drawn from rotation angles in $[-30^{\circ}, 30^{\circ}]$ and translations within $\pm 30$\,mm of the nominal working distance. Frames that failed to provide valid corner clusters were discarded from the calibration optimization but retained in the view count to introduce realistic view redundancy. The overall failure rate was controlled at approximately $0.8\%$. The results, shown in Fig.~\ref{subfig:numofviews}, report the average relative error of the three key intrinsic parameters: the focal length $F$, and the two distance parameters $d_c$ and $d_m$. As the number of views increases, the error decreases and stabilizes: beyond $12$ views, the relative error for $F$ falls below $0.1\%$, while for $d_c$ and $d_m$ it remains below $0.3\%$, confirming the robustness of the method with respect to the number of views.

\textbf{Sensor noise.} Sensor noise was simulated by adding zero-mean Gaussian noise to the raw images, with the noise variance ranging from $0.1$ to $0.5$ in steps of $0.1$ (image intensities normalized to $[0, 1]$). When the variance exceeds $0.5$, the images undergo severe degradation and must be denoised before being input to the calibration pipeline; such cases are therefore excluded. For each noise level, $200$ trials were performed, each using $12$ random-pose light field images. The relative errors of the key parameters are shown in Fig.~\ref{subfig:sensornoise}. When the noise variance is below $0.4$, the error grows approximately linearly with the noise level and remains under $0.75\%$, indicating stable calibration performance. Beyond $0.4$, the error rises sharply as image degradation becomes severe. Notably, the relative error of the focal length $F$ stays below $0.5\%$ across all tested noise levels, demonstrating that the virtual image points $(\hat{q}_x, \hat{q}_y)^T$ estimated by the proposed method remain close to their true positions even under significant sensor noise---a capability that competing light field camera models do not offer.

\textbf{Observation noise.} Observation noise models inaccuracies in the detected feature point coordinates and was simulated by adding zero-mean Gaussian noise with variance ranging from $0.1$ to $1.5$ pixels in steps of $0.1$ to the ground-truth corner positions. For each level, $200$ trials were run. As shown in Fig.~\ref{subfig:cripnoise}, the impact of observation noise on the calibration results is approximately linear. The focal length $F$ again exhibits superior stability, with its relative error never exceeding $0.2\%$, further corroborating the robustness of the virtual point estimation. Moreover, when the observation noise variance is below $0.9$ pixels, the relative errors of all parameters remain under $1\%$, indicating that the calibration method remains robust even in the presence of considerable inaccuracies in corner detection.

\textbf{Comparative analysis.} Comparing the two noise sources reveals that sensor noise has a significantly greater impact on calibration accuracy than observation noise. When the sensor noise variance exceeds $0.4$, the error increases sharply due to image degradation, whereas observation noise does not induce a comparable error jump even at larger variances. The underlying reason is that raw image noise passes through the corner detection algorithm, where its effect is compounded by the detection error itself, resulting in a two-stage accumulation of inaccuracies. In contrast, observation noise applied directly to the corner positions does not undergo this compounding process. Consequently, the calibration system tolerates small fluctuations in corner detection well, but is more sensitive to sensor-level noise. Deploying light field cameras with high-resolution, low-noise sensors is therefore essential for ensuring calibration robustness. In summary, both the full calibration pipeline (including corner detection) and the intrinsic/extrinsic parameter estimation stage alone demonstrate strong robustness under practical noise conditions.

\subsection{Validation on a Self-Built Light Field Camera}

The R12 datasets used in the preceding sections originate from a Galilean-configuration camera with a hexagonal MLA. To assess the generality of the LFT model across different optical designs, we constructed a light field camera that differs in two fundamental aspects: the MLA geometry and the optical configuration. Specifically, we built a focused light field camera (Plenoptic~2.0) employing a square-arranged MLA in the Keplerian configuration, whereas the R12 uses a hexagonal MLA in the Galilean arrangement. Proving that the LFT model works on both architectures is essential, as the hexagonal-to-square shift alters the microlens grid model (Eq.~\ref{eq:grid_final}), and the Galilean-to-Keplerian shift inverts the range of the LFT coefficient $\alpha$ (Eq.~\ref{eq:keplerian}).

\begin{figure}[ht]
    \centering
    \includegraphics[width=0.85\linewidth]{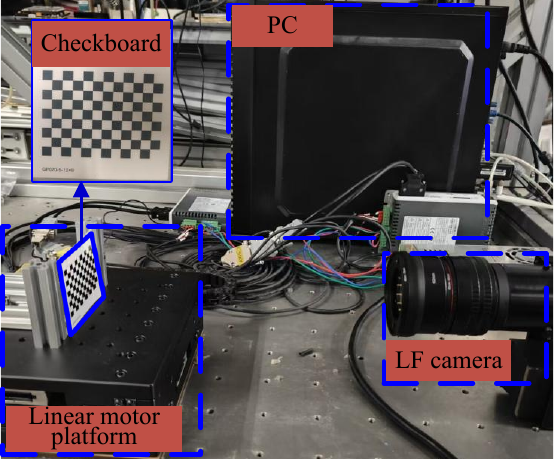} 
    \caption{Experimental platform of the self-built light field camera.}
    \label{fig:selfbuilt_setup}
\end{figure}

The system integrates a Bobcat CLM-B6640 full-frame CCD camera with a KAI-29050 sensor ($6576 \times 4384$ pixels, $5.5$\,\textmu m pixel size), a customized square-arranged MLA, and an exchangeable objective lens. An adjustable clamping mechanism secures the MLA and enables fine alignment. The experimental platform is shown in Fig.~\ref{fig:selfbuilt_setup}. A checkerboard target with $12 \times 9$ corners and a square size of $5$\,mm (manufacturing accuracy $\pm 0.005$\,mm) was used for calibration.

Fourteen free-hand images were acquired at approximately $20$\,cm from the camera. Owing to the higher angular resolution of the self-built system, no frame suffered from the single-corner cluster failure observed in the R12 datasets. Table~\ref{tab:selfbuilt_calib} reports the intrinsic calibration results, where $d_c$ and $d_m$ are converted to the design parameters $a+B$ and $b$ for direct comparison with the physical specifications. The calibrated values agree closely with the design range.

\begin{table}[htbp]
\centering
\caption{Intrinsic calibration of the self-built light field camera.}
\label{tab:selfbuilt_calib}
\begin{tabular}{l c c c c c}
\toprule
 & $F$ (mm) & $a+B$ (mm) & $b$ (mm) & $u_0$ (px) & $v_0$ (px) \\
\midrule
Calibrated & 96.39 & 116.34 & 2.32 & 3299.5 & 2199.4 \\
Design    & 90 & 102--127 & 2.6--4.1 & 3300 & 2200 \\
\bottomrule
\end{tabular}
\end{table}

\begin{figure}[ht]
    \centering
    \includegraphics[width=0.85\linewidth]{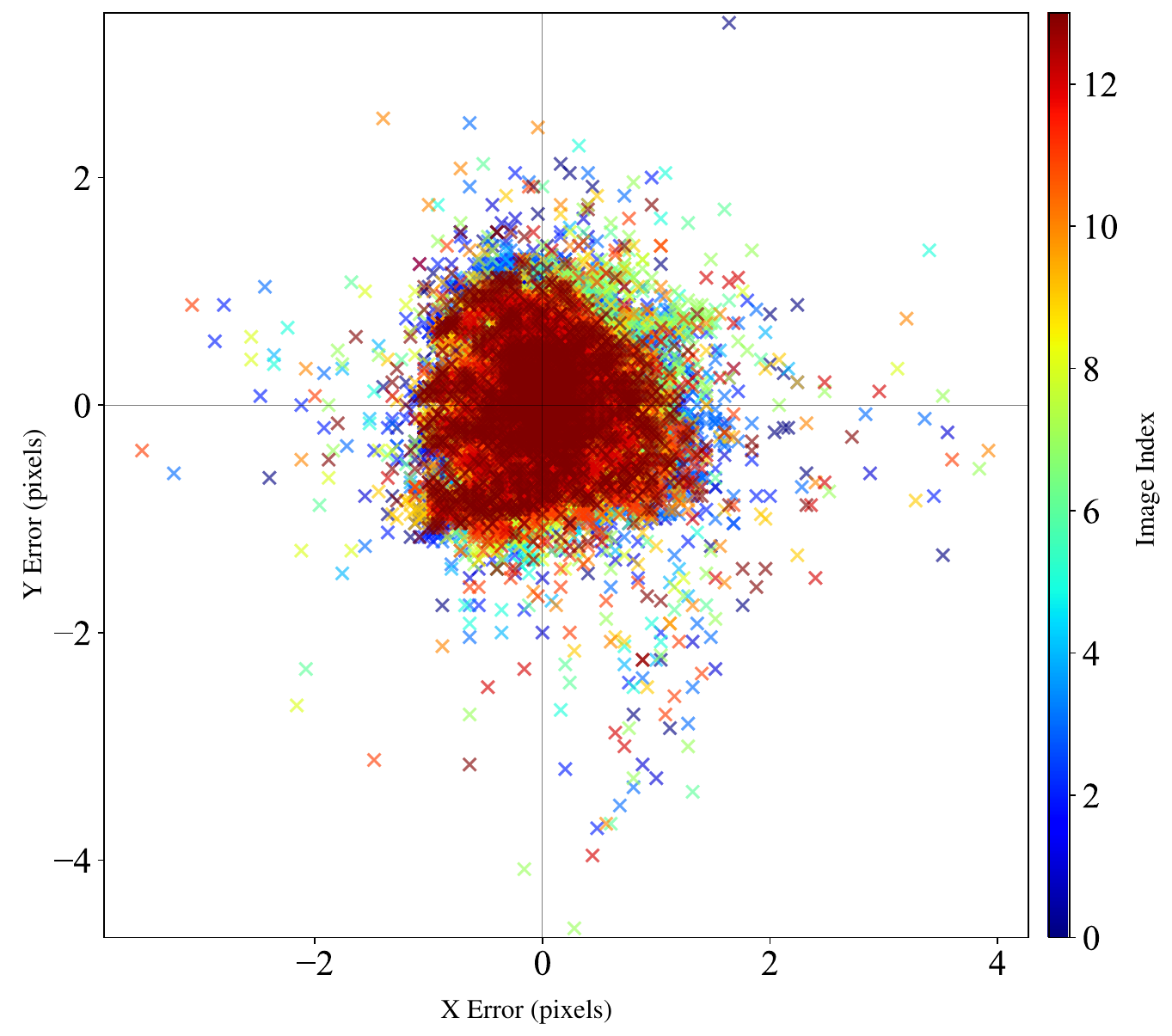}
    \caption{Reprojection error distribution of the self-built camera. Each color represents a different calibration image.}
    \label{fig:selfbuilt_reproj}
\end{figure}

The reprojection error distribution is approximately bivariate normal, with most points falling within $[-1.5, 1.5]$ pixels, as shown in Fig.~\ref{fig:selfbuilt_reproj}. The RMS reprojection error is $0.67$ pixels, confirming the accuracy of both the self-built system and the LFT calibration. All estimated $\alpha$ values exceed $1$, consistent with the Keplerian design and the criterion of Eq.~\ref{eq:keplerian}.

To quantitatively assess the extrinsics, a planar depth estimation experiment was conducted using a high-precision linear motorized stage. Nine images were captured at a fixed interval, with the ground-truth displacement recorded by an optical grating ruler. The mean estimated step size is $9.95$\,mm, yielding a relative error of $0.49\%$ (standard deviation $0.067$\,mm, coefficient of variation $0.67\%$), further corroborating the accuracy of the extrinsic parameter estimation.

\begin{figure}[ht]
    \centering
    \includegraphics[width=0.85\linewidth]{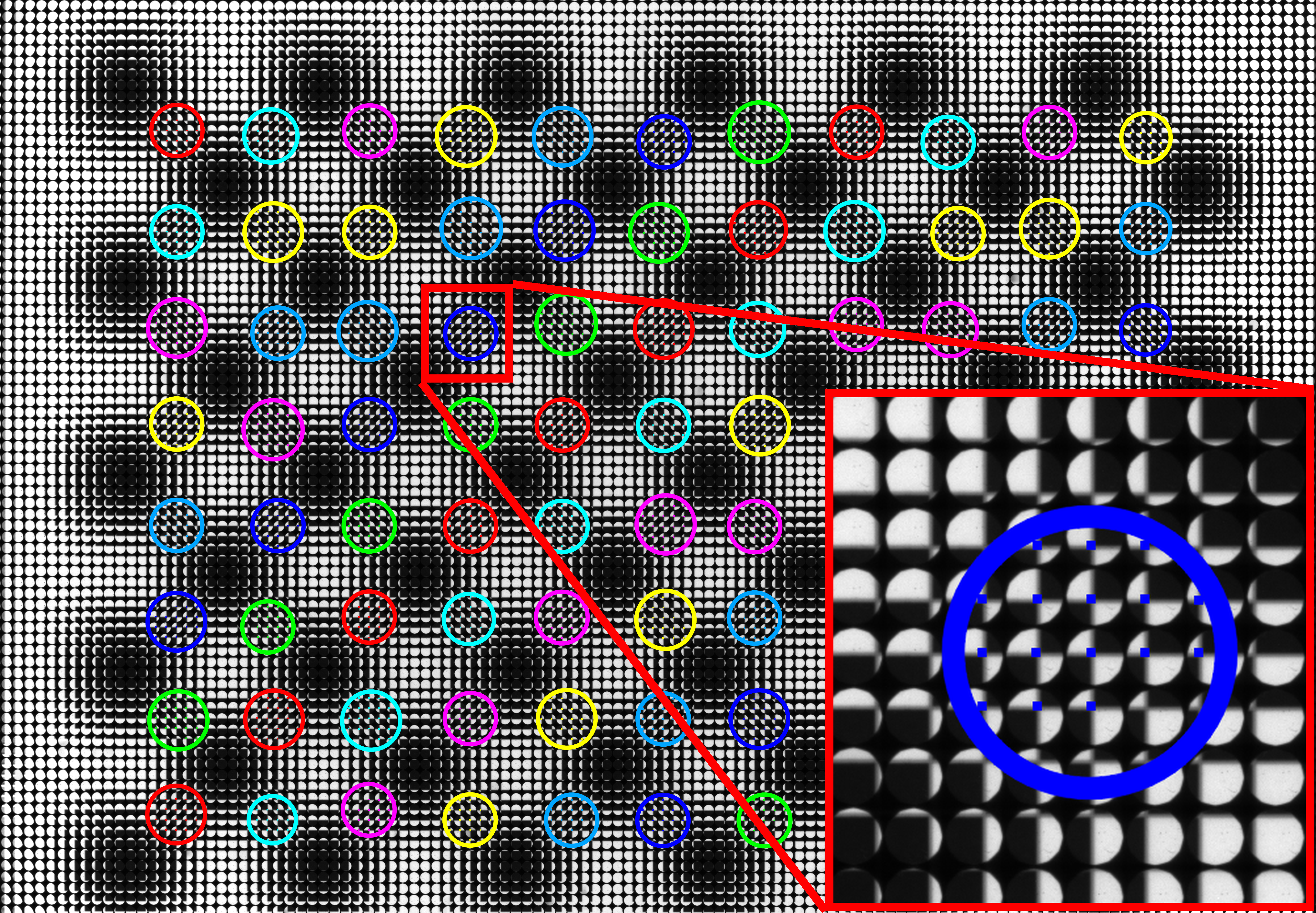}  
    \caption{Corner detection and clustering results on the self-built camera. Different colors represent different clusters.}
    \label{fig:selfbuilt_corners}
\end{figure}

Corner detection on the self-built dataset yields $11 \times 9$ clusters, matching the $12 \times 9$ checkerboard array, as shown in Fig.~\ref{fig:selfbuilt_corners}. Visual inspection confirms that the detected corner positions are in close agreement with the true corners. Overall, these experiments demonstrate that the proposed LFT model and calibration method generalize effectively to custom-built hardware, achieving sub-pixel reprojection accuracy and an extrinsic relative error below $0.5\%$.

\section{Conclusion}

This paper has presented a novel parameter $\alpha$ for light field cameras based on Linear Fractional Transformation (LFT). By decoupling the imaging processes of the main lens and the MLA through a dedicated $\mathbf{H}_\alpha$ matrix, the proposed model enables the main lens and MLA to be calibrated independently, reducing model complexity without sacrificing accuracy. Evaluations on the R12 benchmark datasets demonstrate a mean translation error of $1.38\%$, outperforming the state-of-the-art, while maintaining sub-pixel reprojection accuracy. Validation on a self-built Keplerian light field camera further confirms the practical applicability of the method, achieving an RMS reprojection error of $0.67$ pixels and an extrinsic parameter relative error of $0.49\%$. The complete codebase is openly available to the research community.

The proposed method has several limitations that motivate future work. First, the LFT model requires each object point to be observed by at least two microlenses to form a valid cluster; when this condition is not met, the corresponding frame must be excluded, as observed in two frames of the R12 datasets. Second, the distortion parameters estimated by our pipeline tend to be larger than those reported by competing methods \cite{2022mathieu}, because the least-squares virtual point estimation absorbs some errors that propagate into the distortion model. Future work will explore adaptive threshold selection for corner detection, a light-field-specific distortion model that accounts for the multi-projective nature of MLA imaging, and strategies to recover usable information from frames with fewer than two observations per corner.

\section{Declaration}

This work has been submitted to the IEEE for possible publication. Copyright may be transferred without notice, after which this version may no longer be accessible.

\bibliographystyle{IEEEtran}
\bibliography{ref}

\begin{IEEEbiography}[{\includegraphics[width=1in,height=1.25in,clip,keepaspectratio]{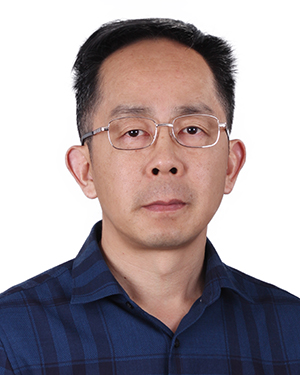}}]{Zhong Chen}
	(Member, IEEE) was born in Chengdu, Sichuan, China, in 1968. He received the B.Eng. degree in mechanical engineering from the Chengdu University of Science and Technology, Chengdu, China, in 1990, the M.Sc. and Ph.D. degrees in mechanical engineering from South China University of Technology, Guangzhou, China, in 1996 and 2003, respectively.
	
	He worked temporarily in Sichuan Chemical factory between 1993 and 1996. From August 15, 2015, to August 14, 2016, he conducted some collaborated researches as a Visiting Scholar in the division of microrobotics and control engineering with Oldenburg University, Germany. He has been holding a Deputy Director with Guangdong provincial key laboratory of precision equipment and manufacturing technology, since 2003. Since 2003, he has been a Professor at South China University of Technology. His research areas cover precision measurement and testing, dynamics of compliant mechanisms, industrial automation, and machine vision and its application.
\end{IEEEbiography}

\begin{IEEEbiography}[{\includegraphics[width=1in,height=1.25in,clip,keepaspectratio]{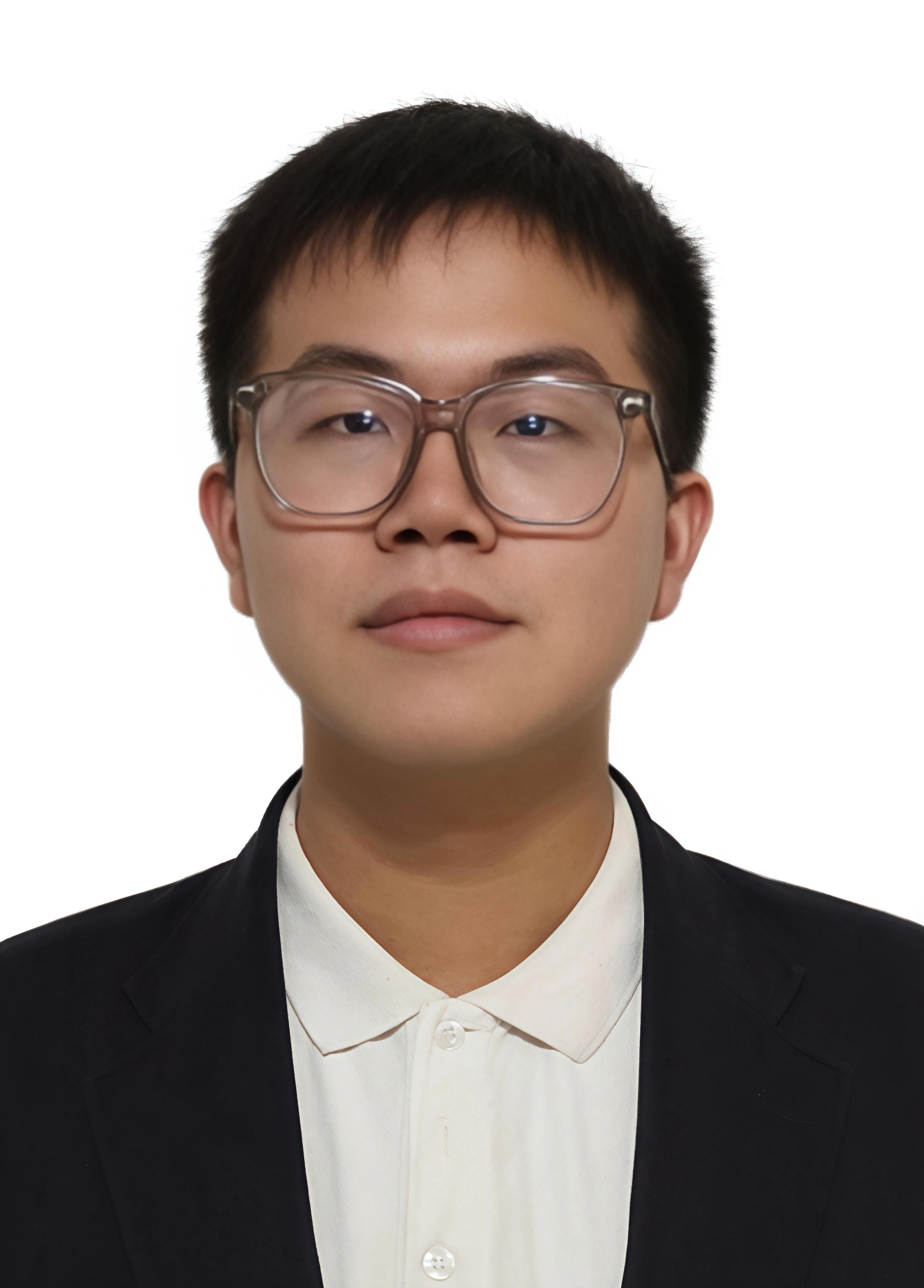}}]{Changfeng Chen} was born in Xinyang, Henan, China, in 2002. He received the B.Eng. degree from Southwest University of Science and Technology, Sichuan, China, in 2023, and is currently working toward the M.SC. degree in mechatronics engineering from South China University of Technology. 
    
    His research interest covers machine vision,3D reconstruction, and plenoptic cameras.
\end{IEEEbiography}

\end{document}